\documentclass[11pt,a4paper]{article}
\usepackage[hyperref]{acl2019}
\usepackage{times}
\usepackage{latexsym}

\usepackage{url}
\usepackage{graphicx, subfig}

\usepackage{amsmath}
\usepackage{amsfonts}
\usepackage{booktabs}
\usepackage{algorithm}
\usepackage{algorithmic}

\aclfinalcopy 
\def\aclpaperid{584} 

\title{Retrieval-Enhanced Adversarial Training for Neural Response Generation}

\author{Qingfu Zhu$^\sharp$, Lei Cui$^\flat$, Weinan Zhang$^\sharp$, Furu Wei$^\flat$ Ting Liu$^\sharp$\thanks{\ \ Corresponding author.} \\
  $^\sharp$Harbin Institute of Technology, Harbin, China \\
  $^\flat$Microsoft Research Asia, Beijing, China\\
  {\tt \{qfzhu, wnzhang, tliu\}@ir.hit.edu.cn} \\
  {\tt \{lecu, fuwei\}@microsoft.com} \\
  }
  
\date{}

\begin{document}
\maketitle

\begin{abstract}
 Dialogue systems are usually built on either generation-based or retrieval-based approaches, 
 yet they do not benefit from the advantages of different models. In this paper, 
 we propose a Retrieval-Enhanced Adversarial Training (REAT) method for neural response generation. 
 Distinct from existing approaches, the REAT method leverages an encoder-decoder framework in terms of an adversarial training paradigm, 
 while taking advantage of N-best response candidates from a retrieval-based system to construct the discriminator. 
 An empirical study on a large scale public available benchmark dataset shows that the REAT method significantly outperforms the vanilla Seq2Seq model as well as the conventional adversarial training approach.
\end{abstract}

\section{Introduction}
Dialogue systems intend to converse with humans with a coherent structure.
They have been widely used in real-world applications, including customer service systems, personal assistants, and chatbots.
Early dialogue systems are often built using the rule-based method \cite{weizenbaum1966eliza} or template-based method~\cite{litman2000njfun,schatzmann2006survey,williams2007partially},
which are usually labor-intensive and difficult to scale up.
Recently, with the rise of social networking, conversational data have accumulated to a considerable scale.
This promoted the development of data-driven methods, including retrieval-based methods~\cite{shang2015neural,sordoni2015seq,vinyals2015seq,Wen2016endtoend} and generation-based methods~\cite{leuski2009building,ji2014information,yan2016learning}.

\begin{table} \small
\centering
\begin{tabular}{ll}
\toprule
MSG & I made strawberry shortcake.\\
\midrule
GT & Where did you learn that, it is sweet and cheery.\\

RSP & How did you {\color{red} \bf make it}? It looks {\color{cyan}\bf delicious}.\\
         
C\#1 & Could you tell me how {\color{red} \bf this thing} {\color{red} \bf is cooked}?\\
           
C\#2 & Tiramisu is my favorite dessert.  It's so {\color{cyan} \bf delicious}.\\
\bottomrule

\end{tabular}
\caption{\label{intro} An example of a message (MSG), a ground-truth response (GT), a generated response (RSP) and N-best response candidates (C\#1 and C\#2) during the training process.
Similar contents in the response and candidates are in boldface.
}
\end{table}

Retrieval-based methods reply to users by searching and re-ranking response candidates from a pre-constructed response set.
Written mainly by humans,
these responses are always diverse and informative, but may be inappropriate to input messages due to their being prepared in advance and thus incapable of being customized~\cite{shang2015neural}.
In contrast, generation-based methods can produce responses tailored to the messages.
The most common method of this category in recent years is the sequence to sequence (Seq2Seq) model~\cite{sutskever2014seq,shang2015neural,vinyals2015seq}.
In practice, it usually suffers from the problem of generating generic responses, such as ``I don't know'' and ``Me, too''~\cite{li2016mmi,serban2016hred}.
While the contents of retrieved responses, apart from the irrelevant parts, are of great diversity, making it a potential resource for tailoring appropriate and informative responses.
Therefore, it is natural to enhance the response generation approach with retrieved responses.

Previous work has been proposed to extend the input of a Seq2Seq model with N-best response candidates (or their contexts)~\cite{song2018ensemble,pandey2018exemplar}.
On one hand, these approaches are trained using MLE objective, which correlates weakly with true quality of responses thus limits the effectiveness of the candidates in producing the responses.
Table~\ref{intro} shows an example during the training process. Related contents of the candidates are appropriately integrated into the response, but the model is discouraged as the response is different from the ground-truth.
On the other hand, rather than just provide materials for the generation, N-best response candidates also contain references for evaluating responses.
Yet they are not efficiently utilized in the objective in the existing training process.

In this paper, we propose a Retrieval-Enhanced Adversarial Training (REAT) approach to make better use of N-best response candidates.
A discriminator is introduced to replace the MLE objective to supervise the training process.
Generated responses containing appropriate and informative contents with input messages are more likely to be seen as human-generated by the discriminator, which encourages the generation model to incorporate more information in candidates into responses.
In addition, the N-best response candidates are also conditioned to the discriminator as references to improve its classification accuracy, which in turn benefits the generation model by adversarial training.
We conduct extensive experiments on a public available NTCIR corpus to verify the effectiveness of the proposed approach, comparing it with retrieval-based methods, generation-based methods, and previous retrieval-enhanced response generation approaches.
The results show that the REAT approach significantly outperforms the baselines in both automatic and human evaluations.

The contributions of this paper are summarized as follows:
\begin{enumerate}
\item We propose a novel retrieval-enhanced neural response generation model adapted from adversarial training approach, which introduces a discriminator to more efficiently utilize the N-best response candidates.
\item Referencing to N-best response candidates, the discriminator of our proposed approach improves over previous discriminators on the classification accuracy.
\item Extensive experiments show that our proposed approach outperforms state-of-the-art baselines in both automatic and human evaluations.

\end{enumerate}

\section{Related Work}
Data-driven dialogue systems can be roughly divided into two categories: retrieval-based and generation-based.
Retrieval-based methods respond to users by selecting the response that best matches an input message from a pre-constructed response set.
\newcite{leuski2009building} match a response with a message using a statistical language model.
\citeauthor{ji2014information}\shortcite{ji2014information} employ information retrieval techniques to rank response candidates.
In addition, the matching and ranking methods can also be implemented using neural networks~\cite{yan2016learning,qiu-EtAl:2017:Short, wu2017sequential}.
Based on that, \newcite{yang2018response} propose a deep matching network which could model external knowledge.

Generation-based methods can be cast as a sequence to sequence (Seq2Seq) process~\cite{shang2015neural,vinyals2015seq,sordoni2015seq} but suffers from generating generic responses.
One way to address the problem is to introduce new content into responses, such as keywords~\cite{mou2016seq2bf,serban2016mrrnn}, topic information~\cite{xing2017topic} and knowledge triples~\cite{zhou2018commonsense}.
Another way is to improve the Seq2Seq architecture.
\citeauthor{li2016mmi}\shortcite{li2016rl} introduce the Maximum Mutual Information as the objective function.
\citeauthor{serban2016vhred}\shortcite{serban2016vhred} add a latent variable to inject variability.
The training of Seq2Seq can be formulated as a reinforcement learning problem~\cite{li2016rl,weinan2017rl}.
To avoid manually defining reward functions, a discriminator can be introduced and trained synchronously by adversarial learning~\cite{li2017adversarial}.
After that, \citet{xu2018diversity} propose a language model based discriminator to better distinguish novel responses from repeated responses.
In a similar adversarial setting, \citet{zhang2018generating} optimize a Variational Information Maximization Objective to improve informativeness.
Our approach is also an adversarial model, the difference is that we employ the N-best response candidates to enhance the generation.

\begin{figure*}[!ht]
\centering
\includegraphics[width=425pt]{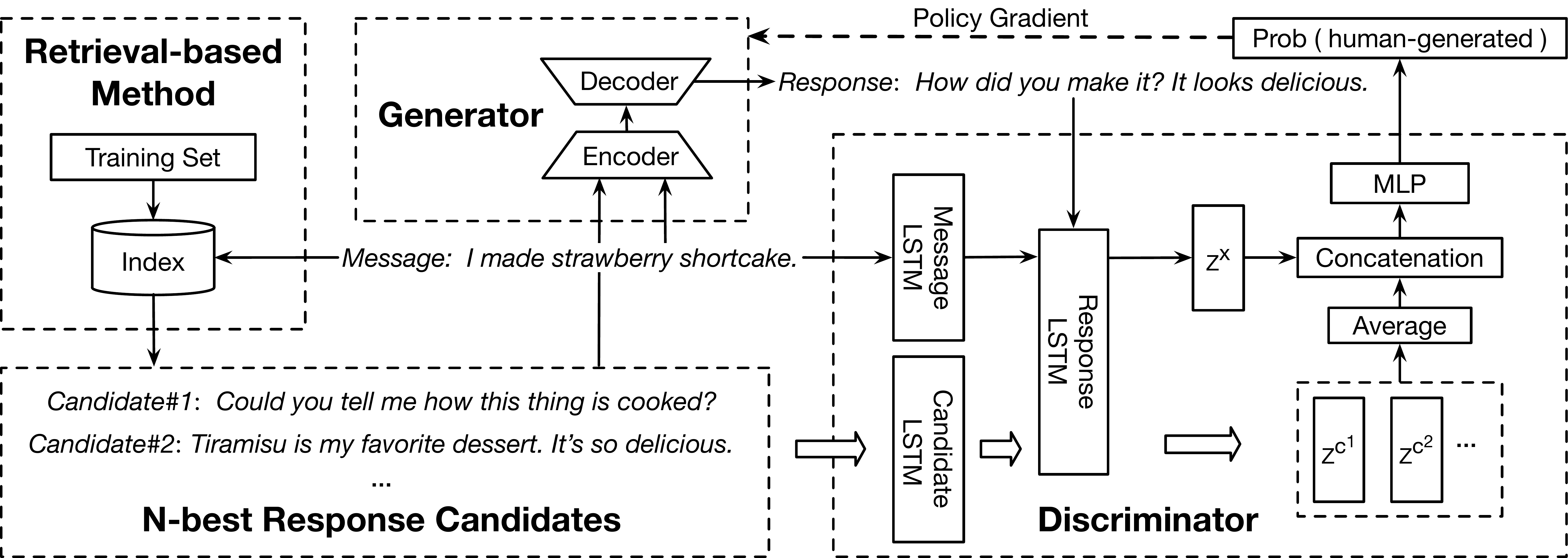}
\caption{An overview of our proposed approach.
The discriminator is enhanced by the N-best response candidates returned by a retrieval-based method.
The discriminator takes as input a response and outputs the probability that the response is human-generated.
The output is then regarded as a reward to guide the generator.
\label{adv}
}
\end{figure*}

Taking advantages of the two methods, retrieval-enhanced response generation approaches make use of the informative content in retrieved results to generate new responses.
Typically, generating responses from retrieved candidates can be seen as a text-to-text system, which produces meaningful text from {\it meaningful text} rather than from abstract meaning representations~\cite{marsi2005explorations}.
\citet{barzilay2005sentence} propose the sentence fusion technique for abstractive multidocument summarization.
In the context of conversation,
\citeauthor{song2018ensemble}\shortcite{song2018ensemble} apply an encoder to every response candidate and integrate the results into the decoding process via the attention mechanism~\cite{bahdanau2014align}.
Similarly, \citeauthor{pandey2018exemplar}\shortcite{pandey2018exemplar} also incorporate response candidates using the attentive encoder-decoder framework on a proposed technical support dataset.
\citeauthor{wu2019response}\shortcite{wu2019response} augments the decoder with an edit vector representing lexical differences between retrieved contexts and the message.
Different from previous work, our approach introduces a discriminator to replace the MLE objective to compute the loss.
Besides, rather than merely being sent to the encoder as generation materials, response candidates in our approach are directly utilized by the discriminator to form a discriminative signal to guide the generator.
The proposed approach is also related to \citeauthor{lin2017adv}\shortcite{lin2017adv}'s work.
They propose an unconditional GAN whose discriminator is augmented with references randomly sampled from the training set for the task of language generation.
In contrast, the proposed approach focuses on the response generation and leverages the message as prior knowledge.
In addition, rather than sampling references from the training set, the candidates in our approach are retrieved according to the relevance to messages using a retrieval-based method.

\section{Method}
In this section, we introduce our proposed REAT approach.
As Figure~\ref{adv} shows,
it consists of two main components: a discriminator $D$ (Sec.~\ref{discriminator}) and a generator $G$ (Sec.~\ref{generator}), both of which are enhanced by N-best response candidates from a retrieval-based method (Sec.~\ref{reference_generation}).
The generator produces a response using the candidates as generation materials.
While in the discriminator, the candidates are provided as references to better distinguish a response, which in turn improves the generator by adversarial training (Sec.~\ref{Adversarial_training}).

\subsection{Retrieval-Enhanced Adversarial Training} \label{Adversarial_training}
The goal of the discriminator is to distinguish whether a response $y$ is human-generated or machine-generated.
It computes the probability $D_{\phi}(y|x, \{c\})$ that the response is human-generated given an input message $x$ and N-best response candidates $\{c\}=\{c^1,... c^k,..., c^N\}$,
where $\phi$ denote the parameters of the discriminator.
Therefore, its objective function is to minimize classification error rate:
\begin{align} \label{dupdate}
J_D(\phi) =& - \mathbb{E}_{y \sim ground-truth}\log D_{\phi}(y|x, \{c\})\nonumber \\
& - \mathbb{E}_{y \sim G} \log (1-D_{\phi}(y|x, \{c\}),
\end{align}

We cast the retrieval-enhanced response generation as a reinforcement learning problem to backpropagate the error computed by the discriminator to the generator via the policy gradient algorithm.
In this way, the generator can be seen as an agent whose parameters $\theta$ define a policy $\pi$.
At each time step, it takes an action $a$ by generating a word and accordingly updates its state $s$, which is defined as a tuple of the message, the candidates and the partially generated response.
At the end of the generation of a response, the agent observes a reward $r$ from the discriminator, which is the probability that the response is human-generated: $D_{\phi}(y|x, \{c\})$.
Here, we do not employ the REGS (reward for every generation step) strategy~\cite{li2017adversarial} as the Monte-Carlo roll-out is quite time-consuming\footnote{Training one epoch takes roughly 120 hours on a TITAN Xp GPU when the roll-out number is 5.} and the accuracy of a discriminator trained on partially decoded sequences is not as good as that trained on complete sequences.

The goal of the agent (the generator) is to minimize the negative expected reward.
With the likelihood ratio trick~\cite{williams1992simple},
the gradient of $\theta$ can be derived as:

\begin{align} \label{gupdate}
J_G(\theta) = & - \mathbb{E}_{y \sim G} (D_{\phi}(y|x, \{c\})),\\
\bigtriangledown J_G(\theta) = &- \mathbb{E}_{y \sim G}( D_{\phi}(y|x, \{c\}) \nonumber \\
& \cdot \bigtriangledown \log G_{\theta}(y|x, \{c\})),
\end{align}
where $G_{\theta}(y|x, \{c\})$ is the probability of generating $y$ given $x$ and $\{c\}$.
In practice, $J_G(\theta)$ and $\bigtriangledown J_G(\theta)$ can be approximated using a single Monte-Carlo sample from $G$~\cite{rennie2017self}:
\begin{align}
    J_G(\theta) \approx & - D_{\phi}(y|x, \{c\}), \quad y \sim G, \\
\bigtriangledown J_G(\theta) \approx & - 
D_{\phi}(y|x, \{c\}) \nonumber \\
        & \cdot \bigtriangledown \log G_{\theta}(y|x, \{c\}), \quad y \sim G. \label{gupdate2}
\end{align}

Both the generator and the discriminator are pre-trained before adversarial training.
The generator is pre-trained on the training set with MLE loss.
The discriminator is pre-trained using human-generated responses as positive samples and machine-generated responses produced by the pre-trained generator as negative samples.

Given the pre-trained generator and discriminator, the adversarial training is a min-max game played between them:
\begin{equation}
\min \limits_{G} \max \limits_{D} J_G(\theta) - J_D(\phi),
\end{equation}
where the discriminator tries to distinguish between human-generated responses and machine-generated responses, while the generator tries to fool the discriminator by producing human-like responses.
The overall algorithm of the retrieval-enhanced adversarial training is summarized as Algorithm~\ref{algo}.

\begin{algorithm}[!t]
\caption{\label{algo}Retrieval-Enhanced Adversarial Training} 
\label{alg:Framwork} 
\begin{algorithmic}[1] 
\REQUIRE ~~\\ 
The training set $\{x, y\}$;
\ENSURE ~~\\ 
The generator parameters $\theta$;\\
The discriminator parameters $\phi$;
\STATE Get N-best response candidates using a retrieval-based method;
\STATE Randomly initialize $\theta$ and $\phi$;
\STATE Pre-train $G$ with MLE loss;
\STATE Generate responses using the pre-trained $G$;
\STATE Pre-train $D$ using machine-generated responses as negative samples and human-generated responses as positive samples;
\FOR{epoch in number of epochs}
\FOR{$g$ in g-steps}
\STATE Update $\theta$ according to Equation \ref{gupdate2};
\ENDFOR
\FOR{$d$ in d-steps}
\STATE Sample $y$ from G as a negative sample;
\STATE Sample $y$ from the human-generated responses as a positive sample;
\STATE Update $\phi$ according to Equation \ref{dupdate};
\ENDFOR
\ENDFOR
\RETURN $\theta, \phi$; 
\end{algorithmic}
\end{algorithm}

\subsection{Discriminator} \label{discriminator}
The discriminator is a binary classifier. 
It takes as input a response $y$, a message $x$, and N-best response candidates $\{c\}$, and subsequently computes a binary probability distribution to indicate whether $y$ is human-generated or machine-generated.

First, we compute a candidate-aware response representation $z^c$ to model the interaction between the candidates and the response.
Each candidate is encoded by a candidate LSTM~\cite{hochreiter1997long}:
\begin{equation} \label{clstm}
u^k_i = f_{c}(c^k_i, u^k_{i-1}), 
\end{equation}
where $c^k_i$ is the $i$-th word of the $k$-th candidate.
$u^k_i$ denotes the hidden state of the candidate LSTM at time step $i$.
$f_c$ is the computation unit of the candidate LSTM.
The initial hidden state $u^k_0$ is set to the zero vector and the last hidden state $u^k_T$ ($T$ denotes the length of an utterence through out the paper) can be seen as a representation of the candidate.
Subsequently, $u^k_T$ is used to initialize the hidden state of a response LSTM, which computes a local candidate-aware response representation $z^{c^k}$  for each candidate $c^k$:
\begin{align} \label{rlstm}
v_{i}^{k} = & f_y(y_i, v^k_{i-1}), \quad z^{c^k} = v^k_{T},
\end{align}
where $v_{i}^{k}$ represents the hidden state of the response LSTM at time step $i$ with regard to the $k$-th candidate.
$f_y$ is the computation unit of the response LSTM and $y_i$ is the $i$-th word of the response.
The candidate-aware response representation $z^c$ is the average of all local candidate-aware response representations: 
\begin{equation}
    z^c = \frac{1}{N} \sum_{k=1}^{N} z^{c^k},
\end{equation}

Second, the interaction between the message and the response is modeled by a message-aware response representation $z^x$ using a message LSTM and the response LSTM introduced above in a similar way to Equation~\ref{clstm} and~\ref{rlstm}.

Finally, the probability that the response is human-generated $D_{\phi}(y|x, \{c\})$ is computed by a Multilayer Perception (MLP):
\begin{equation}
D_{\phi}(y|x, \{c\}) = \sigma(\text{MLP}([z^x, z^c])),
\end{equation}
where the bracket $[\cdot,\cdot]$ denotes concatenation.
$\sigma$ is the sigmoid function\footnote{We did study more complicated relationship among $x$, $y$ ,and $\{c\}$ with  bi-directional LSTM and attention mechanism in the discriminator, but observed no further improvement on the validation set.}.

\subsection{Generator} \label{generator}
The generator \emph{G} is a multi-source Seq2Seq model, which consists of an encoder and a decoder.
The encoder reads from a message and N-best response candidates, summarizing them into context vectors.
The decoder is a language model which produces a response word by word, conditioned with the context vectors.

The encoder first employs a bi-directional LSTM to represent each candidate word and its context information in a response candidate:
\begin{equation} \label{birnn}
\stackrel{\rightarrow}{{h}^k_i} = g^{0}_c (c^k_i, \stackrel{\rightarrow}{{h}^k_{i-1}}), \quad
\stackrel{\leftarrow}{{h}^k_i} = g^1_c(c^k_i, \stackrel{ \leftarrow}{{h}^k_{i+1}}),
\end{equation}
where $g^0_c$ and $g^1_c$ denote the computation units of a forward LSTM and a backward LSTM, respectively.
$\stackrel{\rightarrow}{{h}^k_i}$
and $\stackrel{\leftarrow}{{h}^k_i}$ are the $i$-th hidden states of the two LSTMs.
After that, hidden states in the two directions are concatenated, i.e., $h^k_i = [\stackrel{\rightarrow}{{h}^k_i},\stackrel{\leftarrow}{{h}^k_i}]$.

To capture the different importance of a candidate word in the word-level and the sentence-level, the encoder employs a two-level attention structure.
The word-level attention models the relevance of a candidate word to the decoding context within a candidate, i.e, the word-level attention at the $j$-th decoding time step is computed as:
\begin{equation} \label{attc}
\alpha_{ij}^k = \frac{\text{exp}(q(s_{j-1}, h^k_i))}{\sum_{t=1}^{T}\text{exp}(q(s_{j-1}, h^k_t))},
\end{equation}
where $\alpha_{ij}^k$ is the word-level weight for the $i$-th word of $c^k$.
$s_{j-1}$ is the hidden state of the decoder, representing the decoding context at time step $j$.
$q$ is a feed-forward network.
Considering that different candidates are of different importance, the word-level weights are then rescaled by a sentence-level attention:
\begin{align} \label{weighted_sum}
&a^{c^k}_j = \sum_{i=1}^{T} \alpha_{ij}^k h^k_i, \\
\beta_{kj} =& \frac{\text{exp}(q(s_{j-1}, a^{c^k}_j))}{\sum_{n=1}^{N}\text{exp}(q(s_{j-1}, a^{c^n}_j))}.
\end{align}
where $a^{c^k}_j$ can be seen as a representation of $c^k$.
$\beta_{kj}$ is the sentence-level weight of $c^k$.
The candidate context vector $a^{c}_j$ is then computed taking into account the two-level attention weights:
\begin{equation}
    a^{c}_j = \sum_{k=1}^{N}\sum_{i=1}^{T} \beta_{kj}\alpha_{ij}^k h^k_i
\end{equation}

Meanwhile, the message context vector $a^x_j$ is computed using a message bi-directional LSTM and a word-level attention in a similar way to Equation~\ref{birnn}, \ref{attc} and \ref{weighted_sum}.
Then, the decoder LSTM updates its hidden state conditioned with the context vectors and subsequently generate a word for a response as a standard language model:
\begin{align}
s_j = g_y&([y_{j-1}, a^c_j, a^x_j], s_{j-1}).
\end{align}
where $g_y$ is the computation unit of the decoder.

\subsection{Retrieval-based Method} \label{reference_generation}

To get the N-best response candidates, a retrieval-based method is built using the Lucene\footnote{https://lucene.apache.org/} library and the state-of-the-art response ranking model~\cite{yang2018response}.
First, we merge all message-response pairs whose messages are identical into a document and subsequently build the index for all the documents in the training set.
Second, we use each message as a query to search for $K$ (set to 10) documents whose messages are similar to the query.
After that, responses in the retrieved documents are re-ranked by the ranking model according to their matching scores to the query.
Finally, the top $N$ (set to 2, as in~\citealp{song2018ensemble}) responses are returned as the N-best response candidates.

Note that when we collect N-best response candidates for a training message, the most similar document retrieved is always the one whose message is exactly the training message and responses contain the ground-truth response.
We thus remove the document from the retrieved result before re-ranking to make sure that the N-best response candidates are different from the ground-truth response.

\begin{table}
\centering
\begin{tabular}{lrr}
\toprule
Corpus & \# of message	 & \# of response\\
\midrule
Training    & 119,941 &   1,932,258  \\
Validation  & 10,000    &   163,126  \\
Test        & 10,000    &   162,230  \\

\bottomrule
\end{tabular}
\caption{\label{dataset} Some statistics of the datasets.
}
\end{table}

\section{Experiments}
\subsection{Data}
We use the NTCIR corpus\footnote{http://research.nii.ac.jp/ntcir/data/data-en.html} in our experiments.
Its data are collected from a Chinese microblogging service, Sina Weibo\footnote{https://weibo.com}, where users can both post messages and make comments (responses) on other users' messages.
First, we tokenize each utterance using the Language Technology Platform~\cite{che2010ltp} and remove samples whose responses are shorter than 5, which is helpful in relieving the generic response problem~\cite{li2017adversarial}.
Then, we randomly select 10,000 messages associated with responses to form a validation set and another 10,000 messages with responses as a test set.
Table \ref{dataset} shows some statistics of the datasets.

\begin{table*}[!ht] \small
\begin{center}
\begin{tabular}{l|rrrrr|rrrrr|rrrrr}
\toprule
 & \multicolumn{5}{c|}{Appropriateness}
 & \multicolumn{5}{c|}{Informativeness}
 & \multicolumn{5}{c}{Grammaticality}\\
  & Mean &  +2 & +1 & 0 & $\kappa$ & Mean &  +2 & +1 & 0 & $\kappa$ & Mean & +2 & +1 & 0 & $\kappa$\\ 
\midrule
Rtr  
    & 0.63 & 24.8 & 12.9 & 62.3  & 0.71 
    & 0.92 & 41.1 & 10.1 & 48.8 & 0.67
    & 1.93 & 94.9 & 3.1 & 2.0  & 0.61 \\
    
S2S 
    & 0.76 & 27.9 & 20.0 & 52.1 & 0.58 
    & 0.51 & 10.2 & 30.5 & 59.3  & 0.69 
    & 1.74 & 85.3 & 2.9 & 11.8  & 0.83\\
    
MS2S 
     & 0.85 & 31.9 & 21.5 & 46.6  & 0.63
     & 0.62 & 14.1 & 33.8 & 52.1  & 0.73
     & 1.74 & 85.5 & 3.2 & 11.3  & 0.82 \\
     
Edit  
     & 0.85 & 31.4 & 21.9 & 46.7 & 0.66 
     & 0.67 & 15.9 & 34.9 & 49.2  & 0.68
     & 1.92 & 95.2 & 1.5 & 3.3  & 0.63\\
     
AL 
     & 0.98 & 36.8 & 24.0 & 39.2  & 0.57
     & 0.77 & 21.8 & 33.6 & 44.6  & 0.66
     & 1.88 & 91.7 & 4.7 & 3.6 & 0.58 \\
Ours 
     & 1.10 & 41.5 & 26.8 & 31.7 & 0.65
     & 0.88 & 31.2 & 25.9 & 42.9  & 0.72
     & 1.87 & 89.6 & 7.6 & 2.8  & 0.60 \\
     
\bottomrule
\end{tabular}
\end{center}
\caption{\label{human_eval} Human evaluation results of mean score, proportions of three levels (+2, +1, and 0), and the agreements measured by Fleiss's Kappa in appropriateness, informativeness, and grammaticality.}
\end{table*}

\subsection{Baselines}

\textbf{Rtr}: The retrieval-based method searches the index for response candidates and subsequently returns the one that best matches the message after re-ranking (see Sec.~\ref{reference_generation} for details).

\noindent \textbf{S2S}: The Seq2Seq model with the attention mechanism \cite{bahdanau2014align}.

\noindent \textbf{MS2S}: The ``multi sequence to sequence''~\cite{song2018ensemble} encodes N-best response candidates using $N$ encoders and subsequently incorporates the results into the decoding process by the attention mechanism.

\noindent \textbf{Edit}: The prototype editing model~\cite{wu2019response} augments the decoder with an edit vector representing lexical differences between retrieved contexts and the message.

\noindent \textbf{AL}: The adversarial learning for neural response generation~\cite{li2017adversarial} is also an adversarial method but is not retrieval-enhanced.
Here, we do not employ the REGS (reward for every generation step) setting as the Monte-Carlo roll-out is quite time-consuming and the accuracy of the discriminator trained on partially decoded sequences is not as good as that on complete sequences.

\begin{table}[!t] \small
\centering
\begin{tabular}{lcc}
\toprule
 &  AL & Ours\\
\midrule
Accuracy & 94.01\% & 95.72\% \\

\bottomrule
\end{tabular}
\caption{\label{d_acc} Classification accuracy of discriminators in AL and our approach.
}
\end{table}

\subsection{Experiment Settings}

We use the published code\footnote{https://github.com/MarkWuNLP/ResponseEdit} for Edit and implement other approaches by an open source framework: Open-NMT~\cite{klein2017opennmt}.
The vocabulary table consists of the most frequent 30,000 words, whose 300-dimensional word embeddings are pre-trained on the training set by Word2Vec \footnote{https://code.google.com/archive/p/word2vec/}.
The number of hidden units for all LSTM in our approach is 500.
The batch size is set to 64.

The discriminator and the generator are trained alternately, where the discriminator is optimized for 20 batches, then switch to the generator for 10 batches.
We use ADAM optimizer whose learning rate is initialized to 0.0001.
In the inference process, we generate responses using beam search with beam size set to 5.

\section{Results}

\subsection{Evaluation Metrics}
\paragraph{Human Evaluation}
We randomly sampled 200 messages from the test set to conduct the human evaluation as it is extremely time-consuming.
Five annotators\footnote{All annotators are well-educated students and have Bachelor or higher degree.} are recruited to judge a response from three aspects~\cite{ke2018generating}:
\begin{itemize}
    \item appropriateness: a response is logical and appropriate to its message.
    \item informativeness: a response has meaningful information relevant to its message.
    \item grammaticality: a response is fluent and grammatical.
\end{itemize}
These aspects are evaluated independently.
For each aspect, three levels are assigned to a response with scores from 0 to +2~\cite{shang2015neural}, where 0 represents bad and +2 represents excellent.
The {\it appropriateness} differs from the {\it informativeness} in that the former focuses on the logical relationship between a message and a response, while the latter evaluates the richness of relevant content.

\paragraph{Automatic Evaluation}
We employ {\it Dist-1} and {\it Dist-2}~\cite{li2016mmi} to evaluate the diversity of responses, where Dist-k is the number of distinct $k$-grams normalized by the total number of words of responses.
We also evaluate the {\it Originality} by computing the ratio of responses that do not appear in the training set~\cite{wu2019response}.

To validate the effectiveness of retrieved candidates in enhancing the discriminator, the classification accuracy of the discriminator in AL and our approach is also reported.
Note that the two discriminators after pre-training or adversarial training cannot be compared directly because they are trained by different negative samples produced by different generators.
We thus create a special dataset for this metric where negative samples are generated by a well-trained generator (otherwise, the accuracy will easily reach nearly 100\% as fixed negative samples of low quality are too easy to be distinguished) of AL in advance.

\subsection{Analysis}
The results of the classification accuracy of different discriminators are presented in Table~\ref{d_acc}.
Trained on an identical dataset, our discriminator achieves higher accuracy than the conventional discriminator in AL.
This indicates that the N-best response candidates are helpful for the discriminator in distinguishing between human-generated responses and machine-generated responses, which could in turn benefit the generator in the adversarial training process (discussed later).

\begin{figure}[!t]
\centering
\includegraphics[width=222pt]{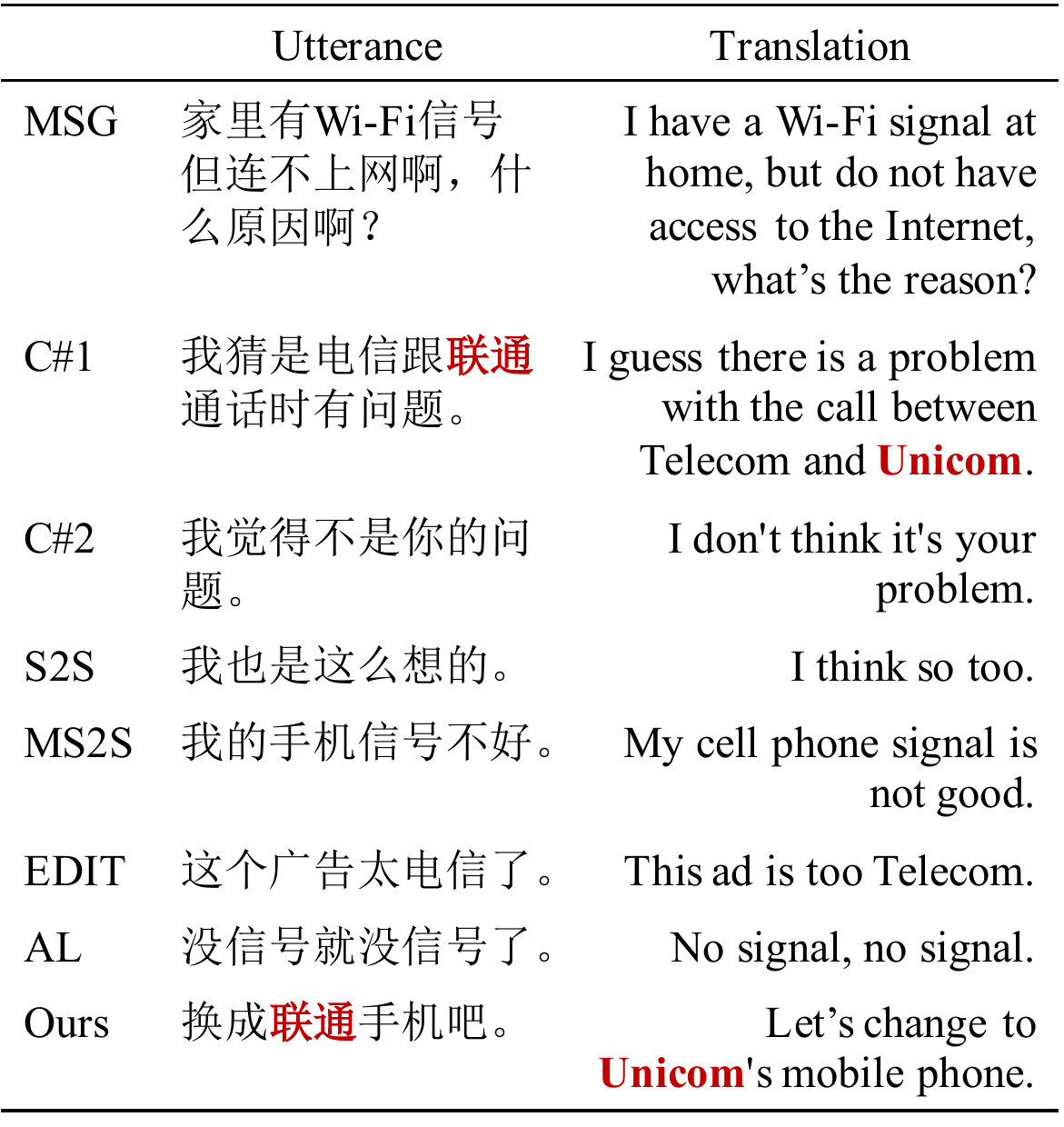}
\caption{\label{samples} An example of a test message (MSG), candidates (C\#1 and C\#2), and responses from different models. The last column are their translations.}
\end{figure}

Table~\ref{human_eval} shows the results of human evaluation.
Our approach has the highest mean score and the largest proportions of +2 and +1 in appropriateness.
Meanwhile, it outperforms all generation-based and retrieval-enhanced approaches in informativeness.
This suggests that our approach is able to respond more appropriately and incorporate informative content into responses at the same time.
Note that Rtr has the highest informativeness mean score due to its diverse human-written content. 
However, it may also contain some irrelevant information, leading to a bad performance in appropriateness.
Besides, most responses in Rtr are annotated as +2 or 0 in informativeness.
This is also because Rtr responses are extremely diverse which always include new content, making a response tend to get +2 if the content are relevant, otherwise 0.
In terms of grammaticality, the mean score of our approach is higher than that of S2S and MS2S, and is comparable with that of AL, indicating that our approach is competitive in generating fluent responses.
Edit has a high mean score mainly due to its relatively simple sentence structure. As shown in Figure~\ref{samples},
S2S and MS2S have similar simple sentence structure to Edit, the reason for the relatively low mean scores of S2S and MS2S in grammaticality is that they have some repetitive responses, like {\it``windows, windows, windows''.}

\begin{table}[t] \small
\centering
\begin{tabular}{lrrrrr}
\toprule
Model & \# of UNI & Dist-1  & \# of BI & Dist-2 & Origin\\
\midrule
Rtr & 6,837 & 0.113 & 25,863 & 0.428 & 0.000\\
S2S  & 1,247 & 0.023 & 3,122  & 0.060 & 0.288\\
MS2S & 2,596 & 0.049 & 6,455  & 0.122 & 0.351\\
EDIT & 1,847 & 0.027 & 5,690  & 0.085 & 0.540\\
AL & 1,760 & 0.033 & 6,697  & 0.124 & 0.590\\
\midrule
D+ & 2,057 & 0.038 & 8,683  & 0.158 & 0.775\\
G+ & 2,440 & 0.046 & 10,461 & 0.200 & 0.792\\
Ours & 3,356 & 0.060 & 13,184 & 0.236 & 0.842\\
\bottomrule
\end{tabular}
\caption{\label{distinct} Automatic evaluation results of the number of distinct uni-grams (\# of UNI) and bi-grams (\# of BI), Dist-1, Dist-2 and Originality (Origin). D+ and G+ are two variants of our approach where candidates are only available for the discriminator and the generator, respectively.}
\end{table}

Agreements among different annotators are calculated by Fleiss' Kappa~\cite{fleiss1971measuring}.
The values of appropriateness and informativeness are all in an interval of (0.4, 0.6] or (0.6, 0.8], which can be seen as ``Moderate agreement'' and ``Substantial agreement'', respectively.
Grammaticality has relatively higher agreement values as it is easier to reach an agreement on 
grammatical errors.

We report the results of Dist-1, Dist-2, and Originality in Table~\ref{distinct}.
AL outperforms S2S in all metrics, indicating that adversarial training is helpful for generating diverse n-grams and responses.
By introducing N-best response candidates, our approach further increases Dist-2 by 0.112 based on AL (from 0.124 to 0.236) and the improvement is significant (t-test, $p<$0.01).
In contrast, the increase of Dist-2 after combining N-best response candidates in MLE based approach is only 0.062, comparing MS2S with S2S.
This suggests that introducing a discriminator with adversarial training is more effective than MLE objective in utilizing N-best response candidates to generate more diverse n-grams.
Note that the improvement after introducing candidates in Dist-1 and Originality is not as significant as that in Dist-2.
This is because responses of MLE based models (MS2S and EDIT) tend to contain informative content with simple sentence structures, like ``... is (not) good.'' (as shown in Figure~\ref{samples}), resulting in high Dist-1 and Originality scores, but their Dist-2 scores are relatively lower than AL and Ours.

To understand the importance of different components of our approach, we also train two variants: D+ and G+, where N-best response candidates are only available for the discriminator and the generator, respectively.
Note that AL does not utilize candidates in the generator nor the discriminator, thus can be seen as a start point of D+ and G+.
As shown in Table~\ref{distinct}, there is an improvement in the performance of both the two variants after introducing the candidates comparing to AL.
The improvement in G+ is more significant as its generator can directly utilize the candidates as generation materials.
While candidates' information in D+ is compressed into a discriminative signal by the discriminator.
Nevertheless, introducing candidates into the discriminator helps to generate more diverse responses comparing AL with D+, and G+ with Ours, demonstrating that the retrieval-enhanced discriminator is able to benefit the generator.

Figure~\ref{samples} shows an example of responses of different models along with the input message and N-best response candidates (C\#1 and C\#2).
The C\#1, which best matches the message among all the candidates, is also the response of the Rtr baseline.
We can see that it contains diverse content, such as ``Unicom'' and ``Telecom''(two telecommunication operators in China, providing broadband, mobile communication as well as customized mobile phones).
However, it talks about ``the call'' between the two operators, which is irrelevant to the message.
The response of S2S is a generic response.
AL has a more diverse response than S2S, however, it does not have access to candidates, which limits the diversity.
MLE based retrieval-enhanced models can make use of the content of candidates, like ``Telecom'' in EDIT, but the way they present the content is not as diverse as ours.

\section{Conclusion and Future Work}
We propose a Retrieval-Enhanced Adversarial Training method for neural response generation in dialogue systems. In contrast to existing approaches, our REAT method directly uses response candidates from retrieval-based systems to improve the discriminator in adversarial training. Therefore, it can benefit from the advantages of retrieval-based response candidates as well as neural responses from generation-based systems. Experiments show that the REAT method significantly improves the quality of the generated responses, which demonstrates the effectiveness of this approach.

In future research, we will further investigate how to better leverage larger training data to improve the REAT method. In addition, we will also explore how to integrate external knowledge in other formats, like the knowledge graph, into adversarial training so that the quality could be further improved.

\section*{Acknowledgments}
The authors would like to thank all the anonymous reviewer for their insightful comments.
The paper is supported by the National Natural Science Foundation of China (No. 61772153).

\bibliography{acl2019}

\begin{thebibliography}{41}
\expandafter\ifx\csname natexlab\endcsname\relax\def\natexlab#1{#1}\fi

\bibitem[{Bahdanau et~al.(2015)Bahdanau, Cho, and Bengio}]{bahdanau2014align}
Dzmitry Bahdanau, Kyunghyun Cho, and Yoshua Bengio. 2015.
\newblock Neural machine translation by jointly learning to align and
  translate.
\newblock In \emph{Proceedings of International Conference on Learning
  Representations}.

\bibitem[{Barzilay and McKeown(2005)}]{barzilay2005sentence}
Regina Barzilay and Kathleen~R McKeown. 2005.
\newblock Sentence fusion for multidocument news summarization.
\newblock \emph{Journal of Computational Linguistics}, 31(3):297--328.

\bibitem[{Che et~al.(2010)Che, Li, and Liu}]{che2010ltp}
Wanxiang Che, Zhenghua Li, and Ting Liu. 2010.
\newblock \href {https://www.aclweb.org/anthology/C10-3004} {{LTP}: A {C}hinese
  language technology platform}.
\newblock In \emph{Coling 2010: Demonstrations}, pages 13--16, Beijing, China.
  Coling 2010 Organizing Committee.

\bibitem[{Fleiss(1971)}]{fleiss1971measuring}
Joseph~L Fleiss. 1971.
\newblock Measuring nominal scale agreement among many raters.
\newblock \emph{Journal of Psychological bulletin}, 76(5):378.

\bibitem[{Hochreiter and Schmidhuber(1997)}]{hochreiter1997long}
Sepp Hochreiter and J{\"u}rgen Schmidhuber. 1997.
\newblock Long short-term memory.
\newblock \emph{Journal of Neural computation}, 9(8):1735--1780.

\bibitem[{Ji et~al.(2014)Ji, Lu, and Li}]{ji2014information}
Zongcheng Ji, Zhengdong Lu, and Hang Li. 2014.
\newblock An information retrieval approach to short text conversation.
\newblock \emph{arXiv preprint arXiv:1408.6988}.

\bibitem[{Ke et~al.(2018)Ke, Guan, Huang, and Zhu}]{ke2018generating}
Pei Ke, Jian Guan, Minlie Huang, and Xiaoyan Zhu. 2018.
\newblock \href {https://www.aclweb.org/anthology/P18-1139} {Generating
  informative responses with controlled sentence function}.
\newblock In \emph{Proceedings of the 56th Annual Meeting of the Association
  for Computational Linguistics (Volume 1: Long Papers)}, pages 1499--1508,
  Melbourne, Australia. Association for Computational Linguistics.

\bibitem[{Klein et~al.(2017)Klein, Kim, Deng, Senellart, and
  Rush}]{klein2017opennmt}
Guillaume Klein, Yoon Kim, Yuntian Deng, Jean Senellart, and Alexander Rush.
  2017.
\newblock \href {https://www.aclweb.org/anthology/P17-4012} {{O}pen{NMT}:
  Open-source toolkit for neural machine translation}.
\newblock In \emph{Proceedings of {ACL} 2017, System Demonstrations}, pages
  67--72, Vancouver, Canada. Association for Computational Linguistics.

\bibitem[{Leuski et~al.(2006)Leuski, Patel, Traum, and
  Kennedy}]{leuski2009building}
Anton Leuski, Ronakkumar Patel, David Traum, and Brandon Kennedy. 2006.
\newblock \href {https://www.aclweb.org/anthology/W06-1303} {Building effective
  question answering characters}.
\newblock In \emph{Proceedings of the 7th {SIG}dial Workshop on Discourse and
  Dialogue}, pages 18--27, Sydney, Australia. Association for Computational
  Linguistics.

\bibitem[{Li et~al.(2016{\natexlab{a}})Li, Galley, Brockett, Gao, and
  Dolan}]{li2016mmi}
Jiwei Li, Michel Galley, Chris Brockett, Jianfeng Gao, and Bill Dolan.
  2016{\natexlab{a}}.
\newblock \href {https://doi.org/10.18653/v1/N16-1014} {A diversity-promoting
  objective function for neural conversation models}.
\newblock In \emph{Proceedings of the 2016 Conference of the North {A}merican
  Chapter of the Association for Computational Linguistics: Human Language
  Technologies}, pages 110--119, San Diego, California. Association for
  Computational Linguistics.

\bibitem[{Li et~al.(2016{\natexlab{b}})Li, Monroe, Ritter, Jurafsky, Galley,
  and Gao}]{li2016rl}
Jiwei Li, Will Monroe, Alan Ritter, Dan Jurafsky, Michel Galley, and Jianfeng
  Gao. 2016{\natexlab{b}}.
\newblock \href {https://doi.org/10.18653/v1/D16-1127} {Deep reinforcement
  learning for dialogue generation}.
\newblock In \emph{Proceedings of the 2016 Conference on Empirical Methods in
  Natural Language Processing}, pages 1192--1202, Austin, Texas. Association
  for Computational Linguistics.

\bibitem[{Li et~al.(2017)Li, Monroe, Shi, Jean, Ritter, and
  Jurafsky}]{li2017adversarial}
Jiwei Li, Will Monroe, Tianlin Shi, S{\'e}bastien Jean, Alan Ritter, and Dan
  Jurafsky. 2017.
\newblock \href {https://doi.org/10.18653/v1/D17-1230} {Adversarial learning
  for neural dialogue generation}.
\newblock In \emph{Proceedings of the 2017 Conference on Empirical Methods in
  Natural Language Processing}, pages 2157--2169, Copenhagen, Denmark.
  Association for Computational Linguistics.

\bibitem[{Lin et~al.(2017)Lin, Li, He, Zhang, and Sun}]{lin2017adv}
Kevin Lin, Dianqi Li, Xiaodong He, Zhengyou Zhang, and Ming-Ting Sun. 2017.
\newblock Adversarial ranking for language generation.
\newblock In \emph{Proceedings of the Thirty-First Conference on Neural
  Information Processing Systems}, pages 3155--3165.

\bibitem[{Litman et~al.(2000)Litman, Singh, Kearns, and
  Walker}]{litman2000njfun}
Diane Litman, Satinder Singh, Michael Kearns, and Marilyn Walker. 2000.
\newblock Njfun: a reinforcement learning spoken dialogue system.
\newblock In \emph{ANLP-NAACL 2000 Workshop: Conversational Systems}, pages
  17--20.

\bibitem[{Marsi and Krahmer(2005)}]{marsi2005explorations}
Erwin Marsi and Emiel Krahmer. 2005.
\newblock Explorations in sentence fusion.
\newblock In \emph{Proceedings of the Tenth European Workshop on Natural
  Language Generation (ENLG-05)}.

\bibitem[{Mou et~al.(2016)Mou, Song, Yan, Li, Zhang, and Jin}]{mou2016seq2bf}
Lili Mou, Yiping Song, Rui Yan, Ge~Li, Lu~Zhang, and Zhi Jin. 2016.
\newblock \href {https://www.aclweb.org/anthology/C16-1316} {Sequence to
  backward and forward sequences: A content-introducing approach to generative
  short-text conversation}.
\newblock In \emph{Proceedings of {COLING} 2016, the 26th International
  Conference on Computational Linguistics: Technical Papers}, pages 3349--3358,
  Osaka, Japan. The COLING 2016 Organizing Committee.

\bibitem[{Pandey et~al.(2018)Pandey, Contractor, Kumar, and
  Joshi}]{pandey2018exemplar}
Gaurav Pandey, Danish Contractor, Vineet Kumar, and Sachindra Joshi. 2018.
\newblock \href {https://www.aclweb.org/anthology/P18-1123} {Exemplar
  encoder-decoder for neural conversation generation}.
\newblock In \emph{Proceedings of the 56th Annual Meeting of the Association
  for Computational Linguistics (Volume 1: Long Papers)}, pages 1329--1338,
  Melbourne, Australia. Association for Computational Linguistics.

\bibitem[{Qiu et~al.(2017)Qiu, Li, Wang, Gao, Chen, Zhao, Chen, Huang, and
  Chu}]{qiu-EtAl:2017:Short}
Minghui Qiu, Feng-Lin Li, Siyu Wang, Xing Gao, Yan Chen, Weipeng Zhao, Haiqing
  Chen, Jun Huang, and Wei Chu. 2017.
\newblock \href {https://doi.org/10.18653/v1/P17-2079} {{A}li{M}e chat: A
  sequence to sequence and rerank based chatbot engine}.
\newblock In \emph{Proceedings of the 55th Annual Meeting of the Association
  for Computational Linguistics (Volume 2: Short Papers)}, pages 498--503,
  Vancouver, Canada. Association for Computational Linguistics.

\bibitem[{Rennie et~al.(2017)Rennie, Marcheret, Mroueh, Ross, and
  Goel}]{rennie2017self}
Steven~J Rennie, Etienne Marcheret, Youssef Mroueh, Jerret Ross, and Vaibhava
  Goel. 2017.
\newblock Self-critical sequence training for image captioning.
\newblock In \emph{Proceedings of the IEEE Conference on Computer Vision and
  Pattern Recognition}, pages 7008--7024.

\bibitem[{Schatzmann et~al.(2006)Schatzmann, Weilhammer, Stuttle, and
  Young}]{schatzmann2006survey}
Jost Schatzmann, Karl Weilhammer, Matt Stuttle, and Steve Young. 2006.
\newblock A survey of statistical user simulation techniques for
  reinforcement-learning of dialogue management strategies.
\newblock \emph{Journal of The Knowledge Engineering Review}, 21(2):97--126.

\bibitem[{Serban et~al.(2016)Serban, Sordoni, Bengio, Courville, and
  Pineau}]{serban2016hred}
Iulian~V Serban, Alessandro Sordoni, Yoshua Bengio, Aaron Courville, and Joelle
  Pineau. 2016.
\newblock Building end-to-end dialogue systems using generative hierarchical
  neural network models.
\newblock In \emph{Proceedings of the Thirtieth AAAI Conference on Artificial
  Intelligence}.

\bibitem[{Serban et~al.(2017{\natexlab{a}})Serban, Klinger, Tesauro,
  Talamadupula, Zhou, Bengio, and Courville}]{serban2016mrrnn}
Iulian~Vlad Serban, Tim Klinger, Gerald Tesauro, Kartik Talamadupula, Bowen
  Zhou, Yoshua Bengio, and Aaron~C Courville. 2017{\natexlab{a}}.
\newblock Multiresolution recurrent neural networks: An application to dialogue
  response generation.
\newblock In \emph{Proceedings of the Thirty-First AAAI Conference on
  Artificial Intelligence}, pages 3288--3294.

\bibitem[{Serban et~al.(2017{\natexlab{b}})Serban, Sordoni, Lowe, Charlin,
  Pineau, Courville, and Bengio}]{serban2016vhred}
Iulian~Vlad Serban, Alessandro Sordoni, Ryan Lowe, Laurent Charlin, Joelle
  Pineau, Aaron Courville, and Yoshua Bengio. 2017{\natexlab{b}}.
\newblock A hierarchical latent variable encoder-decoder model for generating
  dialogues.
\newblock In \emph{Proceedings of the Thirty-First AAAI Conference on
  Artificial Intelligence}.

\bibitem[{Shang et~al.(2015)Shang, Lu, and Li}]{shang2015neural}
Lifeng Shang, Zhengdong Lu, and Hang Li. 2015.
\newblock \href {https://doi.org/10.3115/v1/P15-1152} {Neural responding
  machine for short-text conversation}.
\newblock In \emph{Proceedings of the 53rd Annual Meeting of the Association
  for Computational Linguistics and the 7th International Joint Conference on
  Natural Language Processing (Volume 1: Long Papers)}, pages 1577--1586,
  Beijing, China. Association for Computational Linguistics.

\bibitem[{Song et~al.(2018)Song, Yan, Li, Nie, Zhang, and
  Zhao}]{song2018ensemble}
Yiping Song, Rui Yan, Cheng-Te Li, Jian-Yun Nie, Ming Zhang, and Dongyan Zhao.
  2018.
\newblock An ensemble of retrieval-based and generation-based human-computer
  conversation systems.
\newblock In \emph{Proceedings of the 27th International Joint Conference on
  Artificial Intelligence and the 23rd European Conference on Artificial
  Intelligence}.

\bibitem[{Sordoni et~al.(2015)Sordoni, Galley, Auli, Brockett, Ji, Mitchell,
  Nie, Gao, and Dolan}]{sordoni2015seq}
Alessandro Sordoni, Michel Galley, Michael Auli, Chris Brockett, Yangfeng Ji,
  Margaret Mitchell, Jian-Yun Nie, Jianfeng Gao, and Bill Dolan. 2015.
\newblock \href {https://doi.org/10.3115/v1/N15-1020} {A neural network
  approach to context-sensitive generation of conversational responses}.
\newblock In \emph{Proceedings of the 2015 Conference of the North {A}merican
  Chapter of the Association for Computational Linguistics: Human Language
  Technologies}, pages 196--205, Denver, Colorado. Association for
  Computational Linguistics.

\bibitem[{Sutskever et~al.(2014)Sutskever, Vinyals, and Le}]{sutskever2014seq}
Ilya Sutskever, Oriol Vinyals, and Quoc~V Le. 2014.
\newblock Sequence to sequence learning with neural networks.
\newblock In \emph{Proceedings of the Twenty-Eighth Conference on Neural
  Information Processing Systems}, pages 3104--3112.

\bibitem[{Vinyals and Le(2015)}]{vinyals2015seq}
Oriol Vinyals and Quoc Le. 2015.
\newblock A neural conversational model.
\newblock \emph{arXiv preprint arXiv:1506.05869}.

\bibitem[{Weizenbaum(1966)}]{weizenbaum1966eliza}
Joseph Weizenbaum. 1966.
\newblock Eliza—a computer program for the study of natural language
  communication between man and machine.
\newblock \emph{Journal of Communications of the ACM}, 9(1):36--45.

\bibitem[{Wen et~al.(2017)Wen, Vandyke, Mrk{\v{s}}i{\'c}, Gasic,
  Rojas~Barahona, Su, Ultes, and Young}]{Wen2016endtoend}
Tsung-Hsien Wen, David Vandyke, Nikola Mrk{\v{s}}i{\'c}, Milica Gasic, Lina~M.
  Rojas~Barahona, Pei-Hao Su, Stefan Ultes, and Steve Young. 2017.
\newblock \href {https://www.aclweb.org/anthology/E17-1042} {A network-based
  end-to-end trainable task-oriented dialogue system}.
\newblock In \emph{Proceedings of the 15th Conference of the {E}uropean Chapter
  of the Association for Computational Linguistics: Volume 1, Long Papers},
  pages 438--449, Valencia, Spain. Association for Computational Linguistics.

\bibitem[{Williams and Young(2007)}]{williams2007partially}
Jason~D Williams and Steve Young. 2007.
\newblock Partially observable markov decision processes for spoken dialog
  systems.
\newblock \emph{Journal of Computer Speech \& Language}, 21(2):393--422.

\bibitem[{Williams(1992)}]{williams1992simple}
Ronald~J Williams. 1992.
\newblock Simple statistical gradient-following algorithms for connectionist
  reinforcement learning.
\newblock \emph{Journal of Machine Learning}, 3(8):229--256.

\bibitem[{Wu et~al.(2019)Wu, Wei, Huang, Li, and Zhou}]{wu2019response}
Yu~Wu, Furu Wei, Shaohan Huang, Zhoujun Li, and Ming Zhou. 2019.
\newblock Response generation by context-aware prototype editing.
\newblock In \emph{Proceedings of the Thirty-Third AAAI Conference on
  Artificial Intelligence}.

\bibitem[{Wu et~al.(2017)Wu, Wu, Xing, Xu, Li, and Zhou}]{wu2017sequential}
Yu~Wu, Wei Wu, Chen Xing, Can Xu, Zhoujun Li, and Ming Zhou. 2017.
\newblock A sequential matching framework for multi-turn response selection in
  retrieval-based chatbots.
\newblock \emph{arXiv preprint arXiv:1710.11344}.

\bibitem[{Xing et~al.(2017)Xing, Wu, Wu, Liu, Huang, Zhou, and
  Ma}]{xing2017topic}
Chen Xing, Wei Wu, Yu~Wu, Jie Liu, Yalou Huang, Ming Zhou, and Wei-Ying Ma.
  2017.
\newblock Topic aware neural response generation.
\newblock In \emph{Proceedings of the Thirty-First AAAI Conference on
  Artificial Intelligence}, volume~17, pages 3351--3357.

\bibitem[{Xu et~al.(2018)Xu, Ren, Lin, and Sun}]{xu2018diversity}
Jingjing Xu, Xuancheng Ren, Junyang Lin, and Xu~Sun. 2018.
\newblock \href {https://www.aclweb.org/anthology/D18-1428}
  {Diversity-promoting {GAN}: A cross-entropy based generative adversarial
  network for diversified text generation}.
\newblock In \emph{Proceedings of the 2018 Conference on Empirical Methods in
  Natural Language Processing}, pages 3940--3949, Brussels, Belgium.
  Association for Computational Linguistics.

\bibitem[{Yan et~al.(2016)Yan, Song, and Wu}]{yan2016learning}
Rui Yan, Yiping Song, and Hua Wu. 2016.
\newblock Learning to respond with deep neural networks for retrieval-based
  human-computer conversation system.
\newblock In \emph{Proceedings of the 39th International ACM SIGIR Conference
  on Research and Development in Information Retrieval}, pages 55--64. ACM.

\bibitem[{Yang et~al.(2018)Yang, Qiu, Qu, Guo, Zhang, Croft, Huang, and
  Chen}]{yang2018response}
Liu Yang, Minghui Qiu, Chen Qu, Jiafeng Guo, Yongfeng Zhang, W~Bruce Croft, Jun
  Huang, and Haiqing Chen. 2018.
\newblock Response ranking with deep matching networks and external knowledge
  in information-seeking conversation systems.
\newblock In \emph{Proceedings of The 41st International ACM SIGIR Conference
  on Research and Development in Information Retrieval}, pages 245--254.

\bibitem[{Zhang et~al.(2017)Zhang, Li, Cao, and Liu}]{weinan2017rl}
Weinan Zhang, Lingzhi Li, Dongyan Cao, and Ting Liu. 2017.
\newblock Exploring implicit feedback for open domain conversation generation.
\newblock In \emph{Proceedings of the Thirty-Second AAAI Conference on
  Artificial Intelligence}, pages 547--554.

\bibitem[{Zhang et~al.(2018)Zhang, Galley, Gao, Gan, and
  Dolan}]{zhang2018generating}
Yizhe Zhang, Michel Galley, Jianfeng Gao, Zhe Gan, and Bill Dolan. 2018.
\newblock Generating informative and diverse conversational responses via
  adversarial information maximization.
\newblock In \emph{Proceedings of the Thirty-Second Conference on Neural
  Information Processing Systems}, pages 1810--1820.

\bibitem[{Zhou et~al.(2018)Zhou, Young, Huang, Zhao, Xu, and
  Zhu}]{zhou2018commonsense}
Hao Zhou, Tom Young, Minlie Huang, Haizhou Zhao, Jingfang Xu, and Xiaoyan Zhu.
  2018.
\newblock Commonsense knowledge aware conversation generation with graph
  attention.
\newblock In \emph{the 27th International Joint Conference on Artificial
  Intelligence and the 23rd European Conference on Artificial Intelligence},
  pages 4623--4629.

\end{thebibliography}
\bibliographystyle{acl_natbib}

\appendix

\end{document}